\documentclass[10pt,twocolumn,letterpaper]{article}

\usepackage{iccv}
\usepackage{times}
\usepackage{epsfig}
\usepackage{graphicx}
\usepackage{amsmath}
\usepackage{amssymb}

\usepackage{subfigure}

\usepackage{caption}

\iccvfinalcopy 

\def\iccvPaperID{****} 

\begin{document}

\title{Finding the Topic of a Set of Images}

\author{Gonzalo Vaca-Castano\\
Univeristy of Central Florida\\
{\tt\small gonzalo@knights.ucf.edu}
}

\maketitle

\begin{abstract}
In this paper we introduce the problem of determining the topic that a set of images is describing, where every topic is represented as a set of words. Different from other problems like tag assignment or similar, a) we assume multiple images are used as input instead of single image, b) Input images are typically not visually related, c) Input images are not necessarily semantically close, and d) Output word space is unconstrained. 
In our proposed solution, visual information of each query image is used to retrieve similar images with text labels (tags) from an image database. We consider a scenario where the tags are very noisy and diverse, given that they were obtained by implicit crowd-sourcing in a database of 1 million images and over seventy seven thousand tags. The words or tags associated to each query are processed jointly in a word selection algorithm using random walks that allows to refine the search topic, rejecting words that are not part of the topic and produce a set of words that fairly describe the topic. 
Experiments on a dataset of 300 topics, with up to twenty images per topic, show that our algorithm performs better than the proposed baseline for any number of query images. We also present a new Conditional Random Field (CRF) word mapping algorithm that preserves the semantic similarity of the mapped words, increasing the performance of the results over the baseline.  

\end{abstract}

\section{Introduction}
Images contain lots of information that humans are able to extract in order to offer an interpretation. The interpretation of the image is based on the perception and previous knowledge of the user. Hence, a single image could have multiple interpretations that may lead to ambiguity.  What if instead of having only a single image to be interpreted, we have multiple images that represent the same topic? In that case, we should expect that the information extracted from multiple query images can be leveraged to resolve the ambiguity and have a better description of the set of images.

\begin{figure}
\centering
   \includegraphics[width=\linewidth]{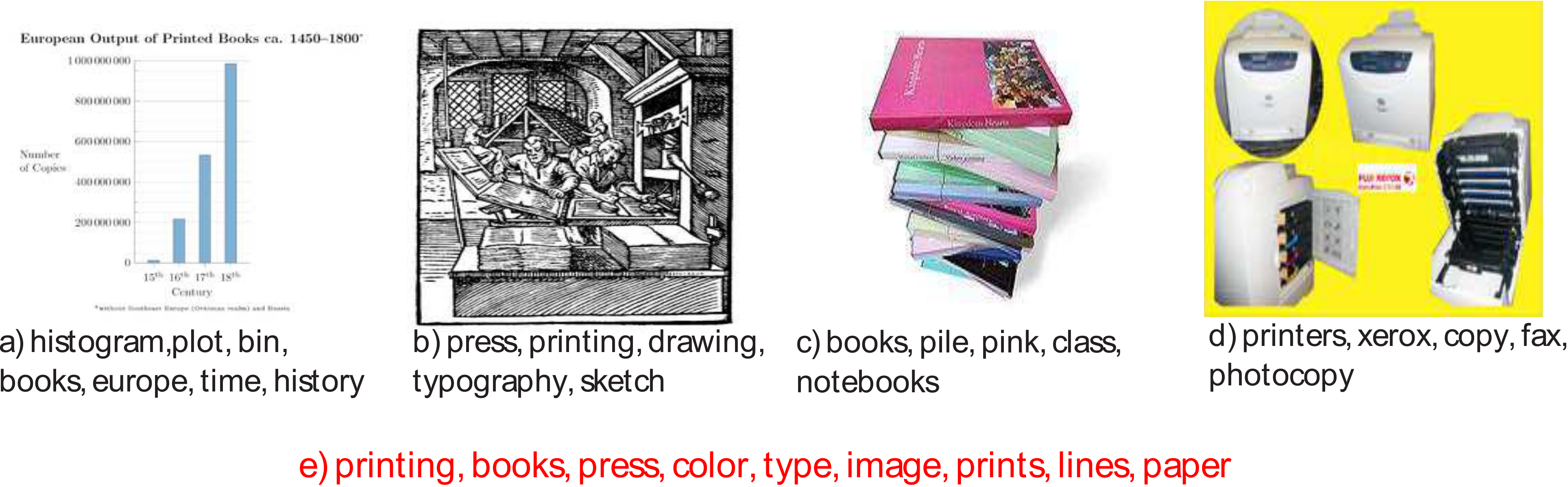}
   \caption{\small{Problem description. Subfigures a) to d) show four different query images and possible text associated to the images. e) shows the expected labels that describe the topic that the images are representing.}}
\label{fig:introduction}
\end{figure}

Following this idea, in this paper we focus on solving the problem of determining the topic that can describe a set of images. Given a set of images that we know for sure belongs to a particular topic, we are interested in finding a set of words that properly describes such topic.

The figure ~\ref{fig:introduction}, from a) to d), shows four possible query images and a set of words that could describe them individually. The obtained set of words are ambiguous. For example, figure a) could be related to histogram or the plot instead of printing process.  Since we know all these images are describing the same topic, our goal is to use the obtained noisy tags of each query image to find a more precise set of words that describe the topic. Text in red e) shows a set of words that properly describes the topic of the four images.

We believe that this type of approach, where multiple related images are used as queries to find the topic that is described, has a lot of potential for several practical applications, where images can be passively captured without direct interaction of the user like in the case of wearable devices and cellphones.  Examples of possible applications include: context discovery, reduction of the search space for object detection, enabling semantic search by improving search accuracy through understanding searcher's intent according to the visual context, video summarization, among others.


The presented problem shares some relation with the auto-tagging problem, but in our case, we assume multiple images are used as input instead of single image. It also differs from others techniques like visual query expansion or other methods that uses multiple queries inputs \cite{Arandjelovic2012a} since in our case, the query images are not visually related. Even more, our input images are not necessarily semantically close.

We also present a solution to the proposed problem. Our solution uses image retrieved from a database that contains noisy words labels, and uses the best retrieved images from each query to generate appropriate  associated word labels. The obtained text representation is processed by a word selection algorithm, with the aim of removing noisy labels obtained from individual query images. The output is a set of words that properly describes the topic of the queried images. Output word space is unconstrained, in the sense that is not manually defined from a set of tags or build from a specific visual domain (scenes, objects, etc), but we also provide an algorithm to map the obtained answers to a more restricted vocabulary. Evaluation is performed in a public dataset containing 300 selected topics with up to 20 images per topic.

Hence, the contributions of our paper are summarized as follows: Firstly, we propose the new problem of determining the topic that a set of images is describing. Likewise, we offered an overall solution to the proposed problem and presented a word selection algorithm that allows to recover words that are conceptually related by means of a random walk using word distances in the word embedding space, and a novel mapping algorithm that allows to map words semantically related to a different vocabulary while preserving semantic relation of the transformed words by means of a Conditional Random Field that exploit distances in the word embedding space to define the pairwise edge potentials. Finally, we proposed an evaluation protocol from a public dataset showing better results compared to the proposed baselines.



\section{Related Work}

The proposed problem of discovering the topic from a set of images, preserve some relation to tag relevance tasks like image tag assignment, and tag reﬁnement, where the outputs are represented as a set of words (tags). Hence, we briefly examine related literature. 
Given an unlabeled image, tag assignment task attempts to assign a (ﬁxed) number of tags related to the image content \cite{Guillaumin2009,Verbeek2010,Chen2013}, while tag refinement deals with  tag ranking of some initial tags from an associated image \cite{Liu2010,Feng2014}. These two closely related problems can be solved by the same type of tag relevance methods, which can be roughly classified in three types\cite{Li2015}: transduction-based, model-based,  and instance-based models. 
Transduction-based methods evaluate tag relevance for a given image-tag pair of a set of images by minimizing some speciﬁc cost function that is typically formulated as a matrix factorization \cite{Zhu2010,Kalayeh2014,Feng2014}. Since the computational complexity involving big matrices, these methods deals with limited number of tags.
Model-based methods are based on parametrized models learned from the training media. Models can be learned over the tags, therefore most of these methods needs to train a classifier such as linear or kernel based SVM\cite{Chena2012} for each tag, making these type of methods hard to scale. Holistic models can also be learned such as Latent Dirichlet Allocation model \cite{Wang2014} in experiments evaluated in 81 tags. 
Instance-based models compares test images with training instances, and therefore is more closely related to our approach. We use similar strategy to the neighbor voting algorithm \cite{Li2009} or variants \cite{Guillaumin2009,Verbeek2010,Chen2013,Pereira2014,Johnson2015} where the relevance of a tag with respect to an image is estimated by counting the occurrence of the tag on the visual neighbors of the annotated samples.  The different variants focus on the type of visual features for retrieval and the different weighting strategies for the retrieved neighbors. In particular, our algorithm uses multiple queries to gather candidate tags, which are refined by promoting tags that are semantically related as measured by distances in the word embedding space. 
The use of multiple images as queries has been relatively unexplored by computer vision researchers. Some authors \cite{Arandjelovic2012a,Basura2013,Chen2012} have focused on image retrieval problem from multiple query images, where query images correspond to instances of the same object under different viewing conditions, and the retrieved images are simply the most similar instances to the queried object. In \cite{Arandjelovic2012a}, text is used as an input to retrieve a set of images of a specific object that are used to retrieve other images of the same object using five different methods. The five methods are based on re-ranking of the images retrieved from a database using an online trained model. To train the model, the set of images initially retrieved is labeled as positive, while a set of randomly selected images of the database is labeled as negative. The authors in \cite{Basura2013} also deal with the problem of object retrieval starting from multiple query images. In their formulation, they derive the most suitable set of patterns to describe the query object, where patterns correspond to local feature configuration. In \cite{Hsiao2015}  the retrieved images are selected using Pareto front method.  In all of the above cases, there is not any notion of semantics or topic selection as in our case since images basically corresponds to different instances of the same class/object. A notable exception is the method presented by Vaca-Castano and Shah \cite{Vaca-Castano2015} where images conceptually related to the query images are retrieved. A drawback of this approach is the difficulty of the evaluation since it relies on user ratings.

\section{Proposed Approach}

\begin{figure*}
\centering
   \includegraphics[height=5cm]{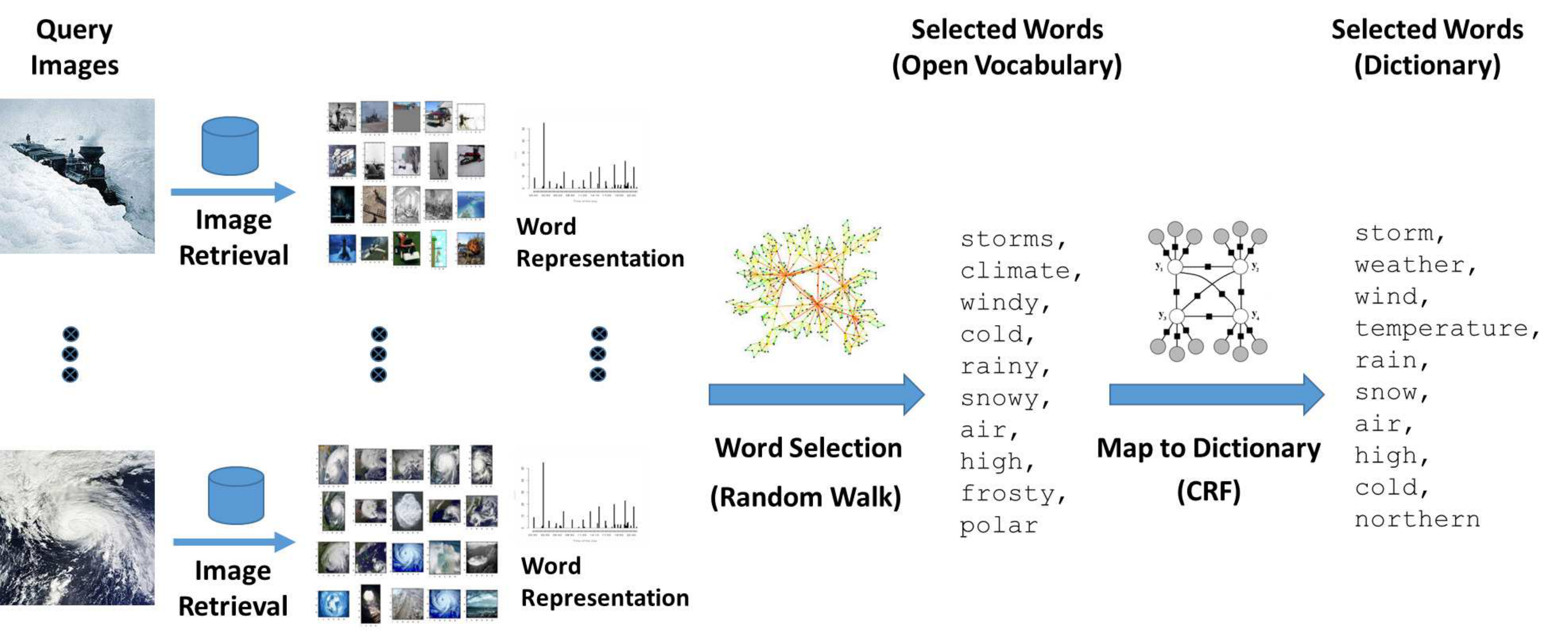}
   \caption{ \small{General Framework of the proposed approach. A database that contains images with weakly labeled text descriptions is used to retrieve images similar to query images. The labels obtained for the retrieved images are used to create a histogram for each query image, where the bins correspond to the frequency of the words in the retrieved set. The sparse representation of each query is used to determine a set of words that best describes the topic of the query images.}}
\label{fig:framework}
\end{figure*}

Figure \ref{fig:framework} presents the general framework of our approach. A database of images is used to retrieve images that are visually similar to each one of the query images. The images in the database contain noisy labels or tags without a defined vocabulary (unbounded number of words) in comparison to other previous approaches for image auto-tagging where the number of tags is limited. The words associated with the retrieved images of a query are weighted and pruned in a selection process that discards words based on two criteria:  How dissimilar is the retrieved image (associated to the word) with respect to the query image, and the frequency of the word or tag among all the retrieved images for a given query. The remaining words are used to generate a histogram that 
produces a joint word representation. A random walk algorithm is performed on the joint histogram representation to rank the words considering their semantic similarity. Finally, a mapping from an open vocabulary to a closed vocabulary using the Conditional Random Field (CRF) algorithm is conducted to select the best set of words, that describes the topic maintaining the semantic coherence among the selected words.

\subsection{Image Retrieval}
Our approach relies on Content-based image retrieval (CBIR) to retrieve images using visual features.   CBIR can be roughly classified into two kinds of algorithms. The first one searches for near-duplicate images either by means of local feature indexing \cite{Nister2006,Jegou2010a,Jain2011,Tolias2013} or hashing of global features like GIST \cite{A.Torralba2008,Oliva2001}. In the the second class of algorithms, the images of the same class are retrieved by using the multiclass classifiers of objects  or attributes \cite{Farhadi2009,F.Yu2012,L.Torresni2010,Douze2011,Wang2011,Zhang2013}. Given the unbounded nature of our problem, supervised classifiers are not considered in our approach. Global features offer advantages in terms of computational complexity, but their performance is lower compared to the local features or combination of them. 

Deep Convolutional Neural Networks (CNN) have attracted the attention of the computer vision community after the success of Krizhevsky \emph{et al.}~\cite{Krizhevsky2012}; pushing up the state-of-the-art in image classification. As was suggested by Krizhevsky, and later confirmed by Babenko \emph{et al.}~\cite{Babenko2014}, the features emerging in the upper layers of the CNN, learned to classify images, serve as good descriptors for image retrieval. Although CNN features were not enough to improve the state of the art for image retrieval, the reported performance is very competitive compared against the local features alternatives that require more time to compute. Hence, we have decided to use CNN as features for image retrieval. We used the Krizhevsky network trained over ImageNet dataset, and removed the last layer of the network of fully connected units. Features are then computed by forward propagation of the mean subtracted and re-scaled to $224 \times 224$ RGB images. The result is a 4,096 dimension vector that represents the image as a global descriptor.
The retrieval results were ordered using Euclidean distance. 


\subsection{Query Word Representation}
An image in the dataset used for Image Retrieval has a dual representation: a visual representation given by a global image descriptor and a textual descriptor represented by a histogram of all of the words describing the image. Every retrieved image has an associated text that serves to link the visual representation with the topic description. Naturally, the vocabulary for the textual description is open and very noisy. Some texts contain mistyped words, words in languages different from English or even texts in other alphabets that are removed; there are also words like pronouns, determiners, and alike that are extremely frequent over the whole dataset, but do not help to identify the topic. A stop list is created to ignore these type of words from the text representation of the dataset.

Retrieved images that are visually closed to the query image should have a higher impact on the text representation of the query images. Hence, the top retrieved candidates are weighted using an inverse exponential function of their visual distance to the query image,
\begin{equation}a_{I_k}=\mathrm{e}^{-\left\|Q - I_k \right\|_2\tau},\label{eq:expdecay}\end{equation}
where $a_{I_k}$ is the weight for the words associated to the image $I_k$,  $\left\|Q - I_k \right\|_2$ is the Euclidean distance between  the visual features of images $I_k$ and query $Q$, and $\tau$ is the exponential time decay constant.

How much weight must be assigned to each retrieved image is influenced by the quality of the retrievals, and is not directly comparable between queries.  We have defined the exponential time decay constant $\tau$ as a linear function of the  mean Euclidean distance of the best top 20 retrieved images of each query. As an effect, the $\tau$ value varies according to the quality of the image retrievals for different queries. 

The textual representation of a query image is computed firstly as the weighted sum of the words that described the retrieved images, where the weight is determined for the exponential function described above followed by performing L1 normalization.  Words that contribute  less than 0.1\% of the total weight  are considered noisy and discarded.

\subsection{Word Selection Algorithm} \label{sec:sel}

Each image query has a representation in terms of words that must jointly be interpreted to select the best set of words that represents the topic. 
There is no way to determine if a query image (and their associated words) has more importance than the other images for  representing a topic; therefore, we assume all the words used to represent query images have the same importance. 
A natural way to choose the best set of words from all the possible candidates is to use  term frequency–-inverse document frequency (tf-idf) representation as a selection criteria. Words with higher weight among candidate words are selected as descriptors of the topic. In our experiments, we use this word selection method as a baseline.  
An inherent limitation of this approach is that every word is considered individually, and the semantic relation that words of a topic must have is not considered.

A word can be represented in a continuous dense vector space that captures semantic knowledge learned in the text domain.  The skip-gram and the Continuous Bag of Words (CBOW) model architectures proposed by Mikolov \emph{et al.} \cite{Mikolov2013a,Mikolov2013b} efficiently learn the semantically-meaningful float point representations of words from very large text datasets. The intuition behind these models is to exploit the fact that given a big corpus of text words, words that are semantically connected tend to appear close to each other in the corpus. The Continuous Bag of Words (CBOW) architecture model has $V$ inputs corresponding to the vocabulary size of the corpus, and the same number of outputs.  Each input is connected to a second layer of dimension D that is shared by all the words of the vocabulary by means of a linear projection. The input projections are averaged and connected to the output layer. The objective is to predict a word given the immediately preceding and following words.  A hierarchical softmax encoded as a Huffman binary tree is used to efficiently save processing.  Each word is a leaf of that tree, and enjoys a path from the root to itself. In this way, N-way normalization of the softmax is replaced by a shorter sequence of $O(logN)$ local (binary) normalizations. All of the text of the corpus is serially used to train this network. Input words that are actives (words around the one that wants to be predicted) are marked as one, and their projection weights are updated doing forward and backward propagation. The resulting projection weights that projects a word into the D-dimensional space becomes the vector representation of the word. The skipgram model is similar to CBOW model, but in this case, an input word is used to predict their previous and following words. A great feature of this model that maps words in vectors is that semantic similarities between two words $W_i$ and $W_j$ can be quantified simply as the cosine distance between the two vector representations,

\begin{equation}d(W_i,W_j)= \frac{W_i \cdot W_j}{\|W_i\|\|W_j\| }.\label{eq:dist}\end{equation}

Word frequency used as word selection criteria for topic description has its flaws. It only looks at the contents of the word, but ignores its influence. All the candidate words from the search are seen as equally important. However, unlike the wrongly selected words, the accurate ones are expected to show a high consistency from a semantic point of view. Therefore, we use Random Walks to score individual words, and discover a reliable subset of words considering the semantic relations between them. 

We define a graph $G=(\textbf{N},E)$ , where $\textbf{N}$ represents the nodes, and $E$ the edges. Each node represents a candidate word $\textbf{N}=\{W_1,W_2,\ldots,W_V\}$. The initial score of the node is given by the frequency of the word using the text representation of the query, and there is an edge between each pairs of nodes $E=\{(W_i,W_j), i\neq j\}$ given by equation \ref{eq:dist} .   

Intuitively, random walks diffuse the score of one node to the neighboring ones if they have a high consistency. This can be imagined by assuming a person is walking from one node of a graph to another and count the number of times each node is visited; the probability of the next node to travel is determined by a predefined consistency between the nodes. If the number of visits to each node is interpreted as a score, after a large number of walks, the nodes which are more consistent to one another will have a higher final score as they are visited more often.

We perform random walks on the constructed graph and update the scores of the nodes using a transition matrix $P$ built from their pairwise similarity given by equation \ref{eq:dist}. As a result, semantically related nodes are encouraged to obtain higher scores. If a random node mistakenly gets a high score due to noisy image retrievals, its score will decrease because of its lack of semantic similarity with other candidate words. On the other hand, if a node is semantically similar to some highly scored node (word), its score will increase after the random walk.

Each random walk iteration will update the scores vector $X$ using the following equation
\begin{equation} X^{t+1}= \alpha P  X^t + (1-\alpha) X^0 ,  \label{eq:random}\end{equation}
where $\alpha$ is a constant between zero and one, and is set to specify the contribution of the initial score versus the pair-wise similarity.
The words with higher scores are selected to represent the topic.

\begin{figure}

\centering
   \includegraphics[height=5cm]{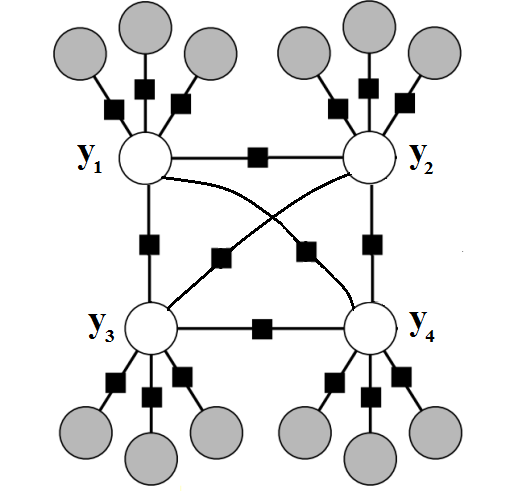}

   \caption{\small{An example of a Graphical Model representation used to map the obtained set of words to a defined dictionary. The figure
shows the scores of the possible word assignments as shaded nodes 
and label assignments  as the white nodes. In this example, the size of the output dictionary $M$ is 3, and the number of words mapped, $L$, is 4.}}
\label{fig:CRFmodel}
\end{figure}

\subsection{Mapping to a Space Spanned by a Dictionary}
The group of words that describes a topic is considered open, in the sense that there is no  limit  on the words that can be used to describe the topic. The employed words are determined by the amount of the text available to describe the images of the retrieval dataset. However, in practical scenarios, it makes sense to limit the words utilized to describe the topics to some smaller set of words. 

Given an output dictionary $\Gamma=\{w_1,w_2,\ldots,w_M\}$, the objective is to find a mapping $\delta:\mathbb O\to\Gamma$  that maps the words from an open vocabulary $\mathbb O=\{o_1 , o_2 , \ldots , o_N\}$  to some words in the dictionary $\Gamma$.  

Mapping a word from an input domain to the target set has to preserve the meaning of the original word. A prospective manner to measure the distances between words from a semantic point was described in section \ref{sec:sel}. We define the mapping $\delta:\mathbb O\to\Gamma$  by selecting for each input word, the closest word from the output dictionary measured as the cosine distance in the vector space projection of the words, as it was defined in equation \ref{eq:dist}. The resulting mapping is considered as our baseline.

The mapping operation is performed on words which are supposed to have certain semantic similarity, as they describe the same topic. The main weakness of the defined mapping is that it totally ignores any relation with the set of words that are going to be mapped.  
Therefore, we propose a Conditional Random Field (CRF) formulation that maps a set of words that have the semantic similarity in another set of words from a reduced dictionary, preserving theirs similarities. 

Let $L$ be the cardinality of the set of words that we want to map, we define a node for each of these $L$ words. Each node has $M$ possible scores corresponding to the distances from the node to each one of the $M$ words of the output dictionary $\Gamma$. The edges between the nodes determine which nodes have the semantic relations. Given that all the words belong to the same topic, these nodes are fully connected.

The figure \ref{fig:CRFmodel} shows the graphical model used in our formulation when a set of four words is mapped and the dictionary size is three words. The white nodes represent the final word label assignments for the set, and the shadowed nodes represent any of the $M$ word possible assignments for the particular node.

Let $Pr(\textbf{y}|G; \lambda)$ be the conditional probability of the word label assignments $\textbf{y}$ given the
graph $G(S_L, Edge)$ and a weight $\lambda$, we need to minimize the energy equation
\small \begin{equation}
−log(Pr(\textbf{y}|G; \omega)) = \sum_{s_i \in S_L}{\psi(y_i|s_i)}+ \lambda \sum_{s_i,s_j \in Edge}{\phi(y_i , y_j |s_i , s_j )},
\end{equation}
\normalsize
where $\psi$ are the unary potentials, and $\phi$ are the pairwise edge potentials.
In our problem the unary potential is computed from $S_i$, the cosine distance between input word and a word in the output dictionary $w_i$,  as 
\begin{equation}
\psi(i)=1-S_i ,
\end{equation}
which privileges word labels with high similarity to the original word. 

The pairwise edge potential is given by a matrix, $V(y_p,y_q)$, that determines the distances between the words that belongs to the dictionary $\Gamma$ as measured by relation $1-d(W_i,W_j)$. The matrix attempts to penalize words that are not related semantically (assigning a penalty), enforcing the global similarity of the labeled words. 

The energy function to minimize can be represented as: 
\begin{equation}
E(\textbf{y})=\sum_{p=1\cdots N}{\psi(p,y_p)} + \sum_{p=1\cdots N, q=1\cdots N} {\lambda_{p,q} V(y_p,y_q)} ,
\label{eq:CRF}
\end{equation}
where $\lambda_{p,q}$ is a weighted adjacency matrix, with weights equal to $1/L$.

We use the graph-cuts based minimization method in ~\cite{Boykov2004,Boykov2001,Kolmogorov2004} to obtain the optimal solution for equation \ref{eq:CRF}.

\begin{figure*}
\centering
   \includegraphics[width=\linewidth]{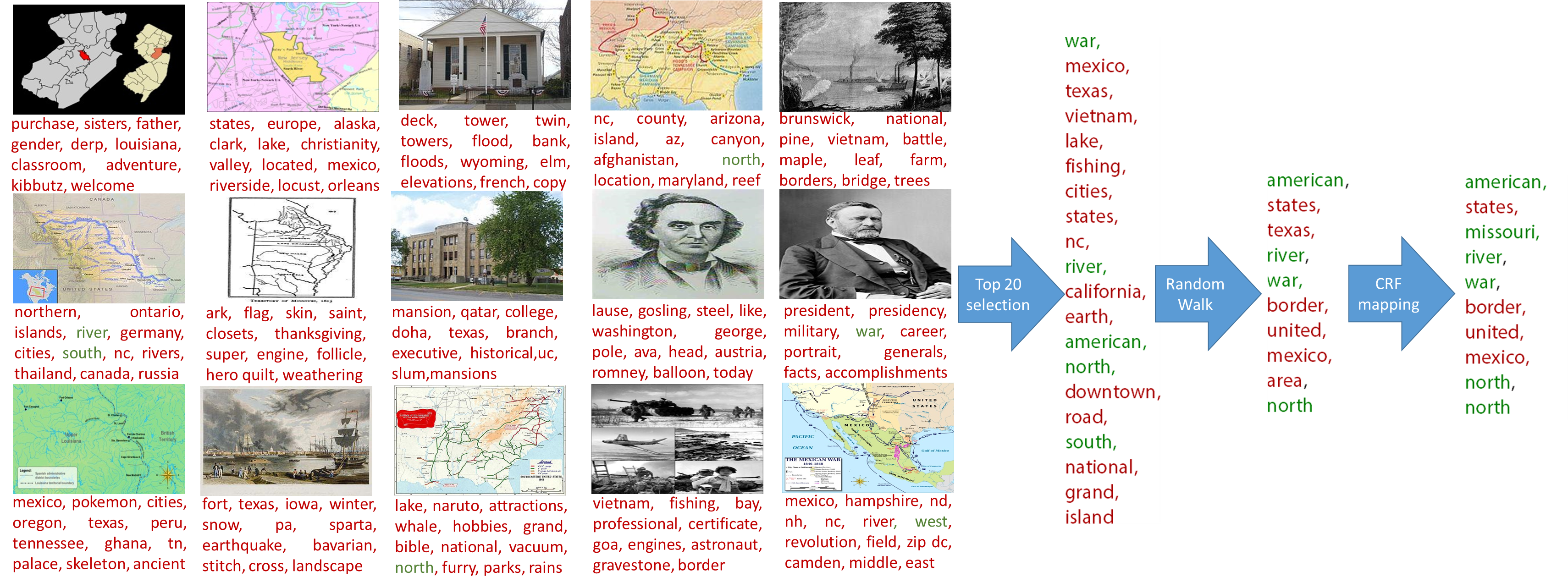}
   \caption{\small{An example of qualitiative results obtained by  the proposed approach.  Words written in \textcolor[rgb]{0.24,0.7,0.44}{green} indicate words that match with the groundtruth, while words written in \textcolor[rgb]{0.75,0,0.13}{red} indicate the words that do not match with any word in the groundtruth. The fifteen images on the left are used as query images to search for the topic. Each query image includes a text showing the top retrieved words associated to the query image. The top 20 highest scored words of the joint representation are shown next. After that, re-ranking is performed using random walk, removing words with low semantic coherence like ``vietnam'', ``earth'', and ``island''. Finally, a mapping to a closed vocabulary is conducted which allows us to transform the word ``texas'' to ``missouri''. }}
\label{fig:qualitative}
\end{figure*}

\section{Experiments and Results}
\subsection {Implementation Details}
In the image retrieval section of our experiments, we use the dataset provided for the 2013 MSR-Bing Image Retrieval Challenge  \cite{Bing2013}, which was sampled from one-year click logs of the Microsoft Bing image search engine.  It consists of 1 million images. All of the images have associated one or more entries in a table with 23 million triads: $< image \,K\,, query \, Q \, , click \,count \,C >$ , which means the image $K$ was clicked $C$ times in the search results of query $Q$ in one year (possibly by different users at different times). A total of 11.7 million different query terms $Q$ are available, and the number of clicks $C$ is always more than zero.

The dataset is very suitable for image retrieval since the texts associated to the images are fairly accurate, because they are built from user's search criteria and their click preferences.  The most important advantage is that the image labeling does not require humans dedicated to this activity, and is the product of implicit crowdsourcing.   The average number of queries per image is 23.1; consequently there is a significant amount of text describing each image. We used this dataset as our image retrieval  dataset.

Our text representation of the images from the dataset, is a histogram of the words of all texts associated to an image (related by the 23 million triads). After removing words in languages different from English, words written in alphabets different from Modern Latin alphabet, and removing pronouns, determiners and extremely frequent words, we end up with a vocabulary of 77,488 words for the Bing Image Dataset.

We use the public domain implementation from caffe library \cite{Jia2013}  and the provided AlexNet trained model to compute the CNN visual features that represent the images of the database. We use as descriptor the $4,096$ dimension vector obtained by running forward the network and removing the last layer with 1,000 units. Visual feature extraction  is performed very efficiently; image resizing and feature extraction of the 1 million images of the database can be performed in less than 24 hours on a regular Quad core personal computer. 
In order to have a fast approximated nearest neighbor retrieval implementation, we use Locality Sensitive Hashing (LSH), which is an algorithm for solving the approximate Near Neighbor Search in high dimensional spaces. LSH hashes input items so that similar items map to the same ``buckets'' with a high probability. The distance computation is only performed within the elements of the same ``bucket'' , increasing the retrieval speed.
In all of our experiments we use 10 bits to generate the ``buckets''.

%

Once the nearest neighbors of the query image are computed, we weight each text representation of the nearest neighbors with the exponential decay function described in equation \ref{eq:expdecay}. Experimentally we found that a good value for $\tau$ is given by $\tau=\mathbb{E}_{1\cdots20}[\left\|Q - I_k \right\|_2]/3$ , where $\mathbb{E}_{1\cdots20}[\left\|Q - I_k \right\|_2]$ is the mean distance of the 20 nearest neighbors to the query in the visual space. 

Finally, the edges of the graph for a random walk, are computed using the equation \ref{eq:dist}, the mapping between any word and its vector representation is given by the projection matrix trained on a part of Google News dataset which contains about 100 billion words\footnote{https://code.google.com/p/word2vec/}. The vocabulary size $D$ of this model is 300.


\subsection{Evaluation}
\subsubsection{Dataset}
We use the dataset \footnote{We are not related with the authors of the dataset} provided in \cite{Aletras2013} \footnote{ http://staffwww.dcs.shef.ac.uk/people/N.Aletras/resources.html} to evaluate our experiments. The dataset consist of 300 topics represented by the top 10 most representative words for the topic. The topics  were found by performing Latent Dirichlet Allocation LDA in a corpus of the New York Times published between May and December 2010, and by randomly selecting Wikipedia categories from a hierarchy in a breadth-first-search manner starting from a few seed categories (e.g. sports, politics, computing)  that have more than 80 articles. 

A set of 20 images is provided for each topic.  These images correspond to the top 20 images under the Creative Commons license from English Wikipedia, retrieved from a search of the top-5 terms from a topic using Google Search.

Additionally, the dataset provide human scores that judge how appropriate the image was as a representation of the main subject of the topic. The score allows us to rank which of the twenty images are more representative of the topic according to human criteria. As we will see later, the order in which the query images are presented have an influence on the quality of the topic description. 

\subsubsection{Jaccard Index as a Performance Metric}
Given a set of query images of the same topic used as input, the result is a set of words that describe the topic. The ground truth for one topic of our experiments is the set of ten words that describes the topic. Hence, for each topic, we need to compare the similarity of two sets of words. The Jaccard index, also known as the Jaccard similarity coefficient, is chosen as metric for the evaluation of our algorithm. The Jaccard index measures similarity between finite sample sets, and is defined as the size of the intersection divided by the size of the union of the sample sets:
\begin{equation} J(G,O)= \frac{|G \cap O |}{|G \cup O|} , \end{equation}
where $G$ is the set of groundtruth words that describes the topic, and $O$ is the output set of words found by the algorithm.


\subsection{Experiments}

The figure \ref{fig:qualitative} presents an example of the proposed approach. In this case, fifteen images are used as query. After performing the image retrieval, and weighting the words associated, we end up with a huge set of words to describe each query image. In the figure, only the top scored words are shown.  Most of these words are written in red, indicating they are not part of the groundtruth for the topic. After combining them, we obtain five matching words among the top 20 of retrieved words.  However, the retrieved list of words contains some elements, which are semantically incoherent like ``vietnam'' or ``fishing''. The random walk algorithm takes care of unrelated words, and also increases the number of matching words in the top 10 to four. Finally, in the last step, we complete a mapping of the found words to a smaller dictionary. In all of our experiments, we used a dictionary of 1,653 words, created by concatenating all the groundtruth words available in the list of 300 topics.  As a result of the mapping step, the word ``texas'' is relabeled as ``missouri'', increasing the number of correctly matched words to five, while other retrieved word like ``area''  is mapped to ``north'' that was already found as topic descriptor, decreasing the number of misleading words.

\begin{figure*}
\centering

\subfigure[Input Images are not sorted] {
	\includegraphics[width=0.32\textwidth]{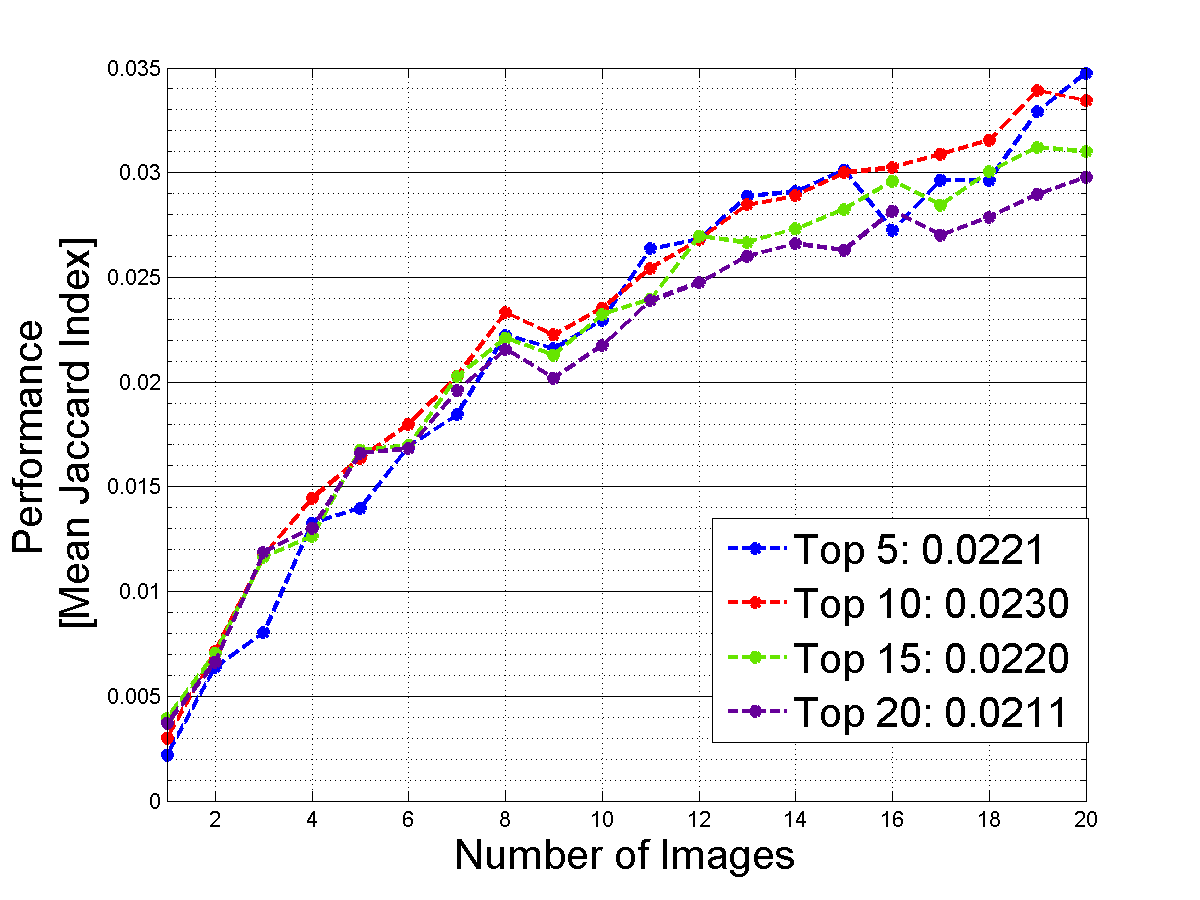} 
}
\subfigure[Most descriptive images are sorted first] {
	\includegraphics[width=0.32\textwidth]{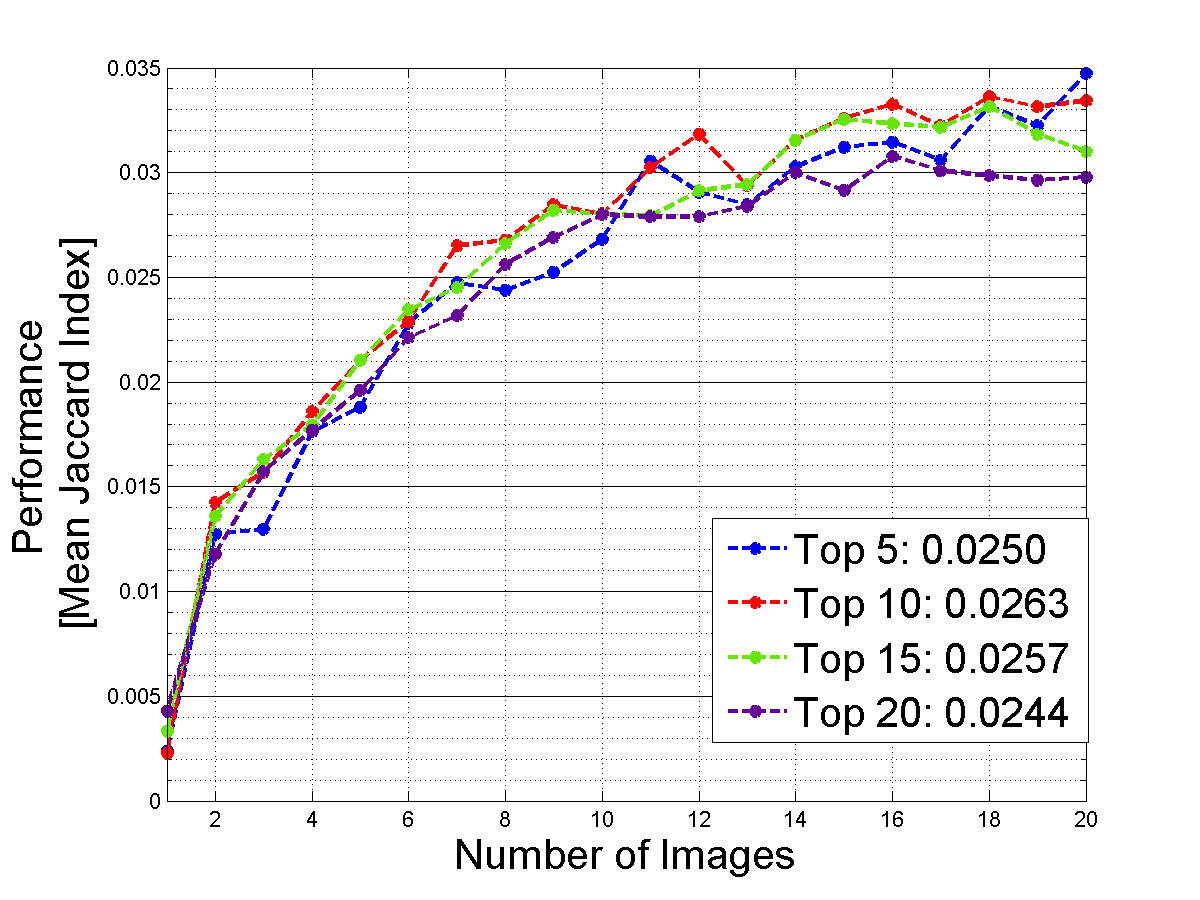} 
}
\subfigure[Less descriptive images are sorted first] {
	\includegraphics[width=0.32\textwidth]{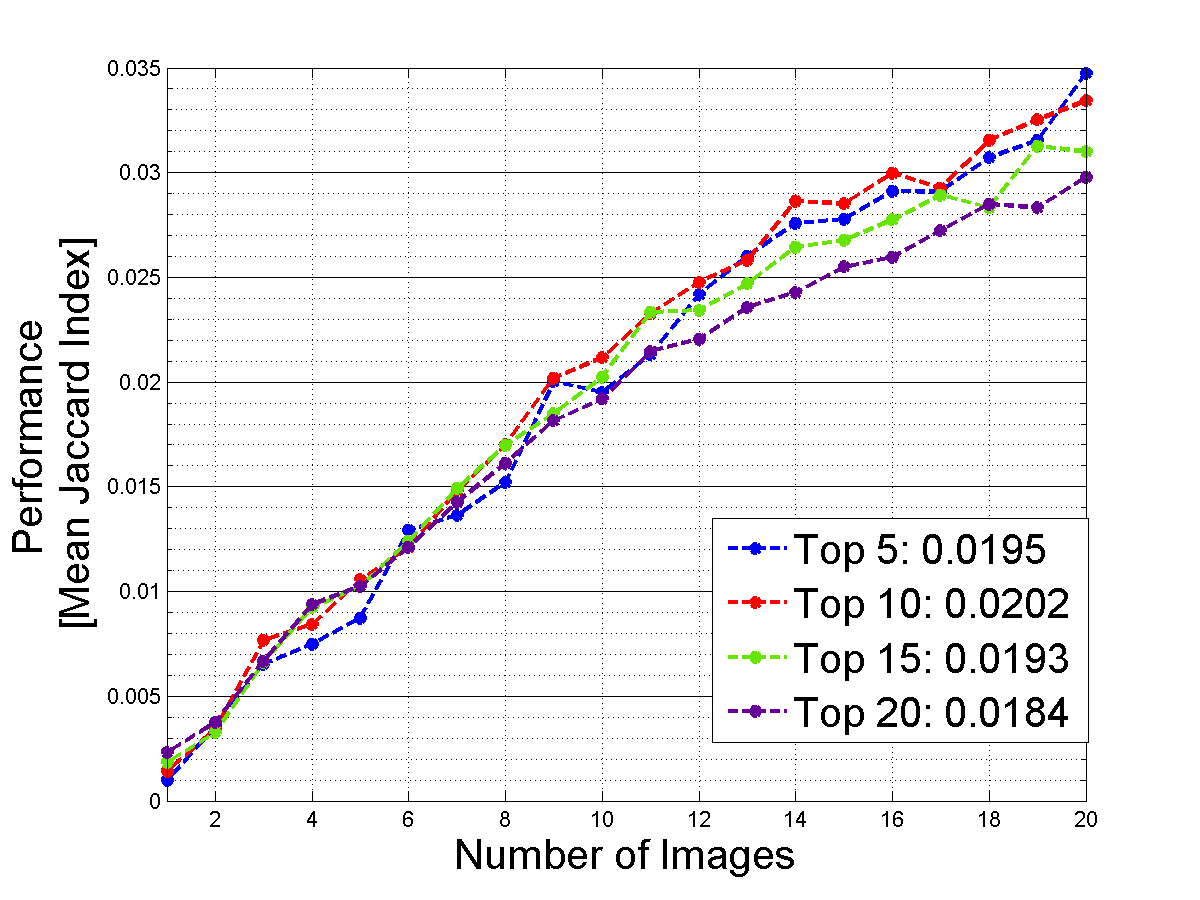} 
}
  
   \caption{\small{Results of the baseline method when the top 5, top 10 , top 15, and top 20 retrieved words are used to represent the topic. The displayed plots show  the performance (Y axis) computed as the mean Jaccard index of the 300 topics, as a function of the number of images used as query. The three plots represent different sorting of the input images, according to how representative the images are for the topic. Retrieving top 10 words produce the best results for any type of sorting. } }
\label{fig:base}
\end{figure*}

\begin{figure*}
\centering
\subfigure[Input Images are not sorted] {
	\includegraphics[width=0.32\textwidth]{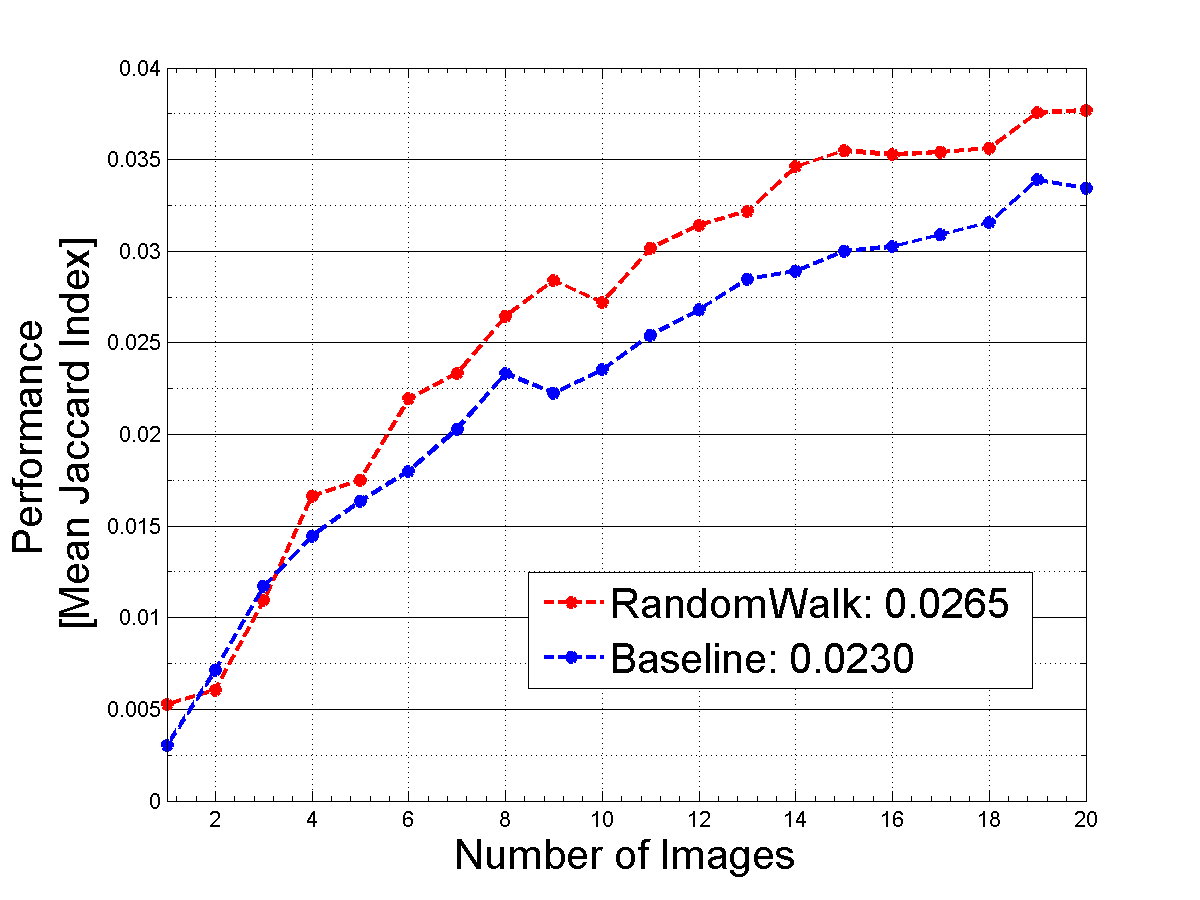} 
}
\subfigure[Most descriptive images are sorted first] {
	\includegraphics[width=0.32\textwidth]{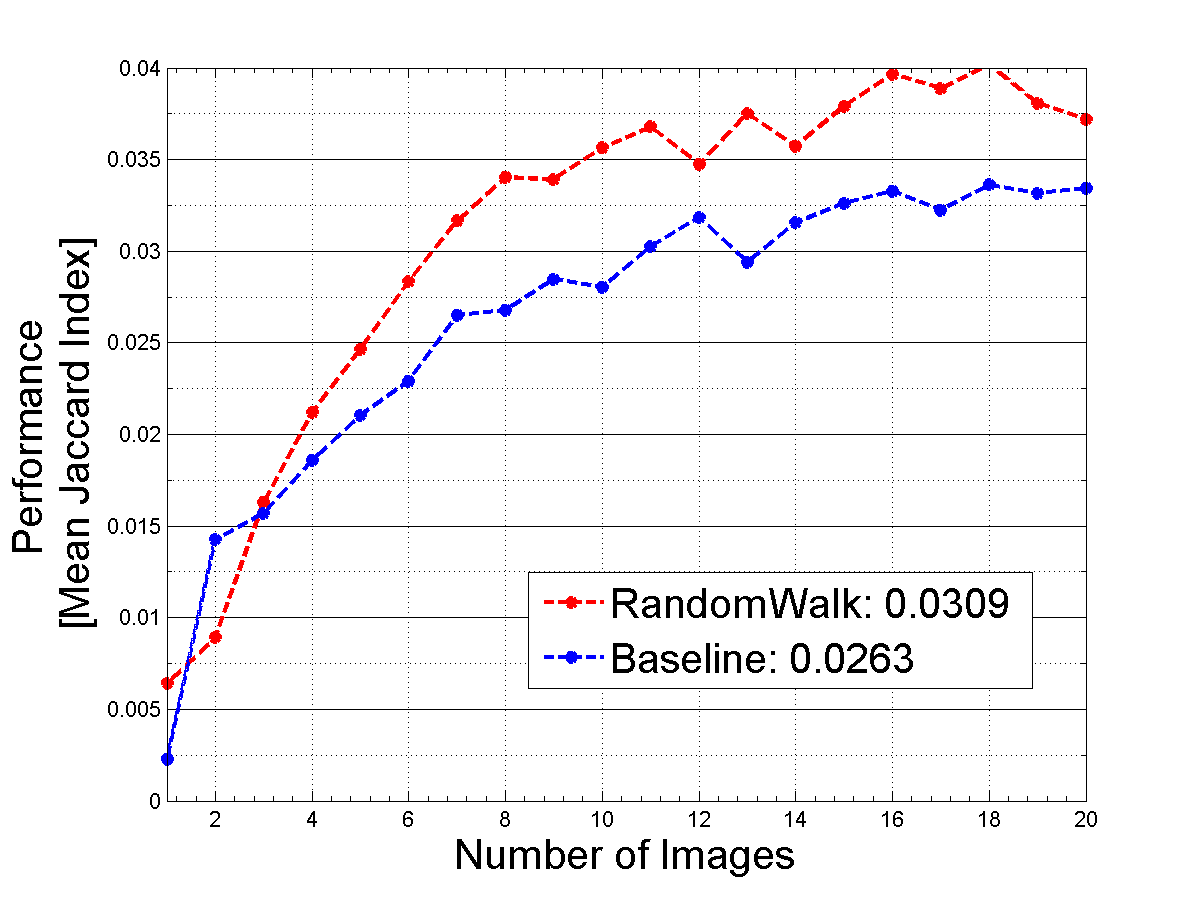} 
}
\subfigure[Less descriptive images are sorted first] {
	\includegraphics[width=0.32\textwidth]{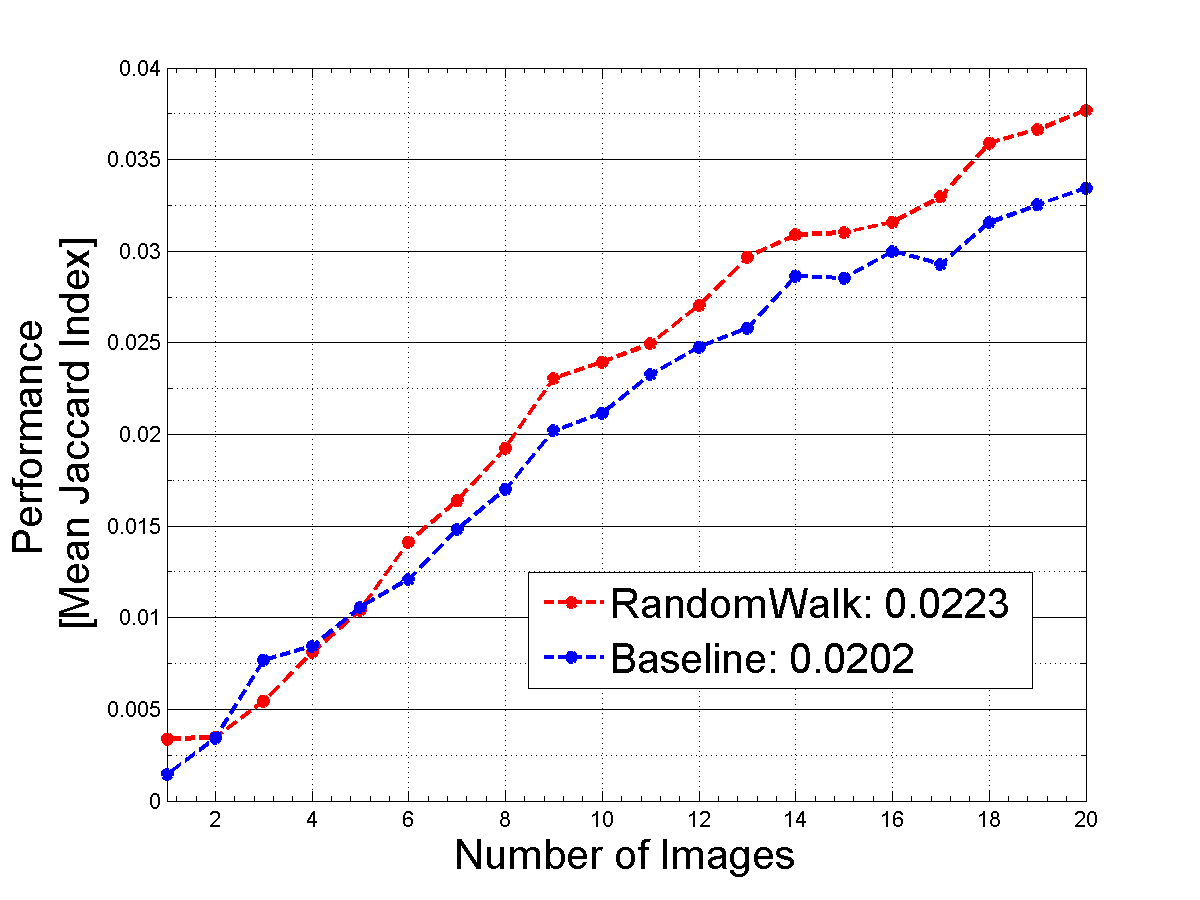} 
}

   \caption{\small{Comparison between the baseline (using top 10 retrieved words) and our random walk algorithm to select the best words that describe the topic.  The three plots represent different sorting of the input image based on how representative the images are for the topic. The random walk algorithm significantly outperforms the baseline under any type of sorting, and any number of input images. }}
\label{fig:base2}
\end{figure*}

\begin{figure*}

\centering
\subfigure[Input Images are not sorted] {
	\includegraphics[width=0.32\textwidth]{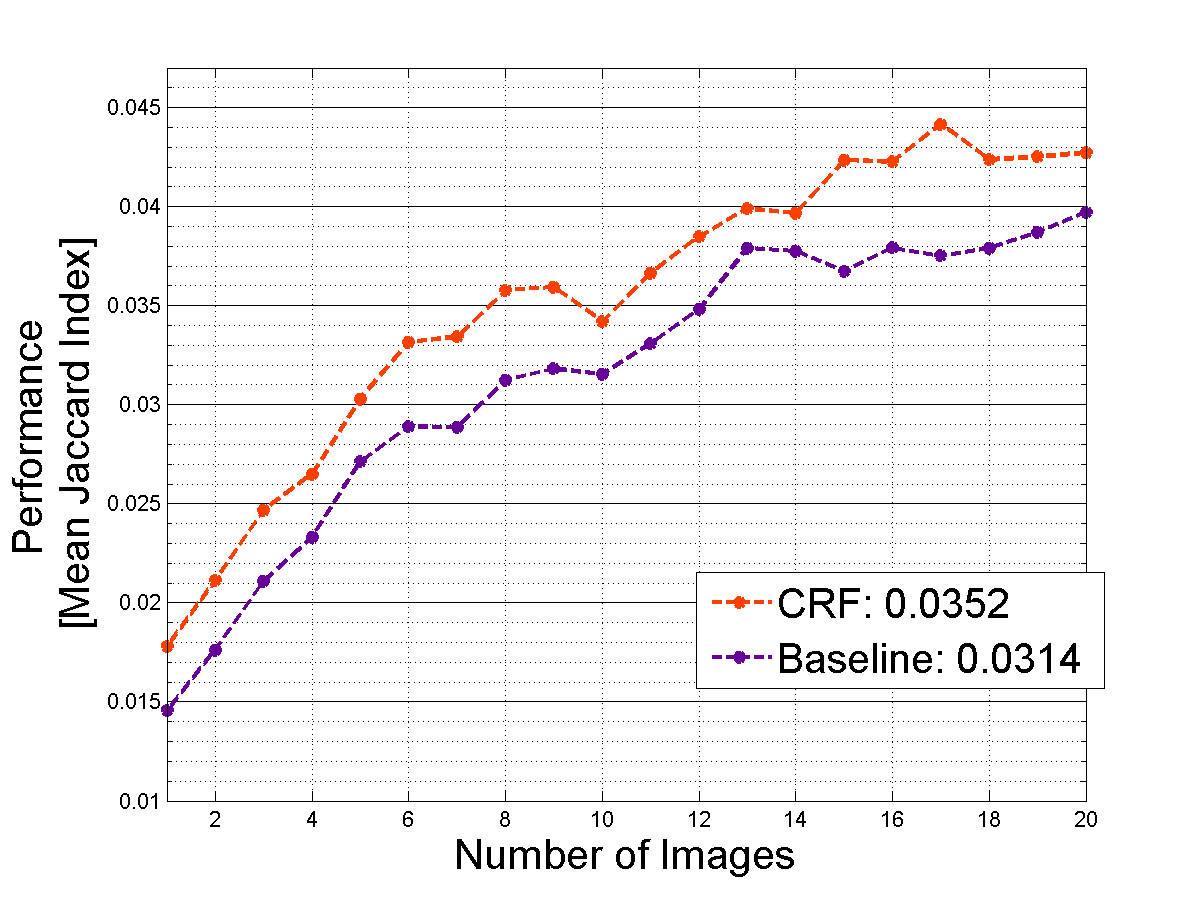} 
}
\subfigure[Most descriptive images are sorted first] {
	\includegraphics[width=0.32\textwidth]{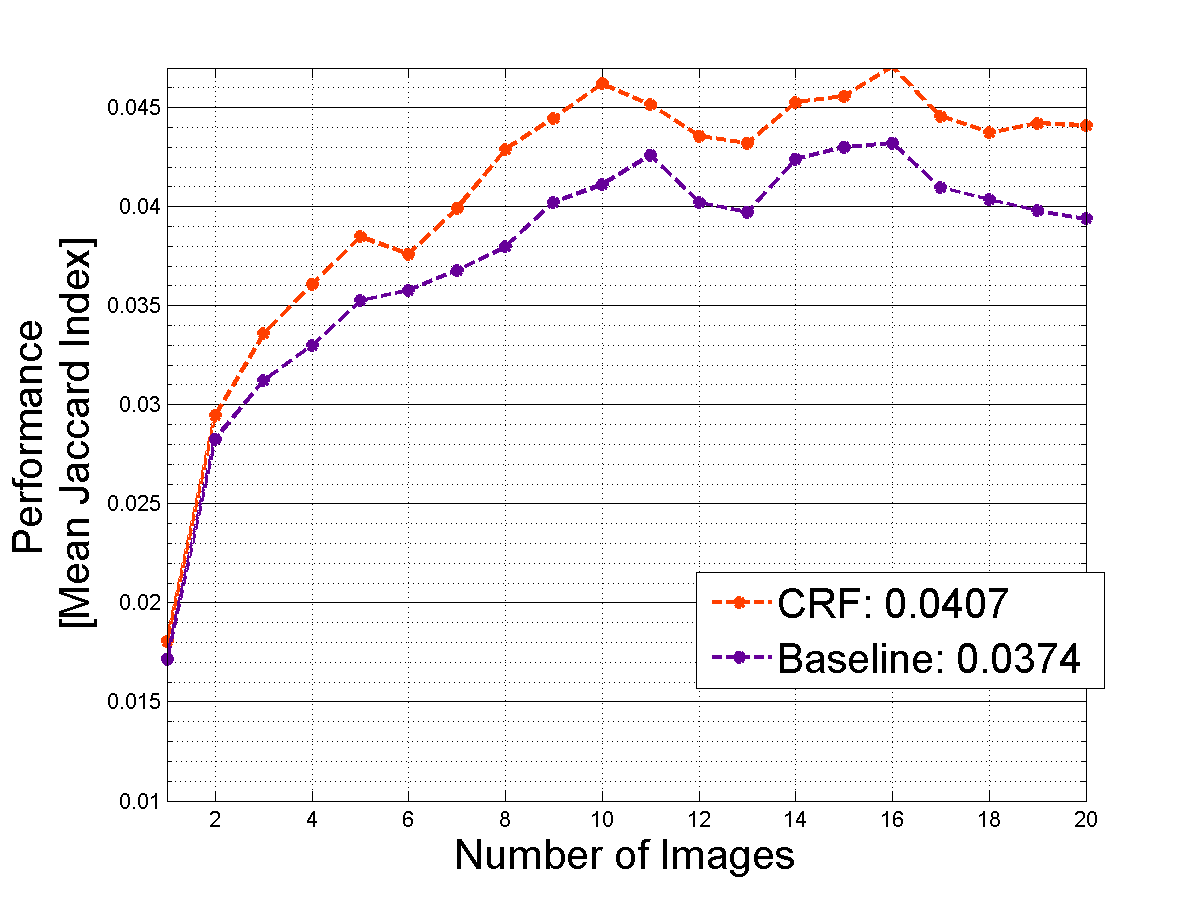} 
}
\subfigure[Less descriptive images are sorted first] {
	\includegraphics[width=0.32\textwidth]{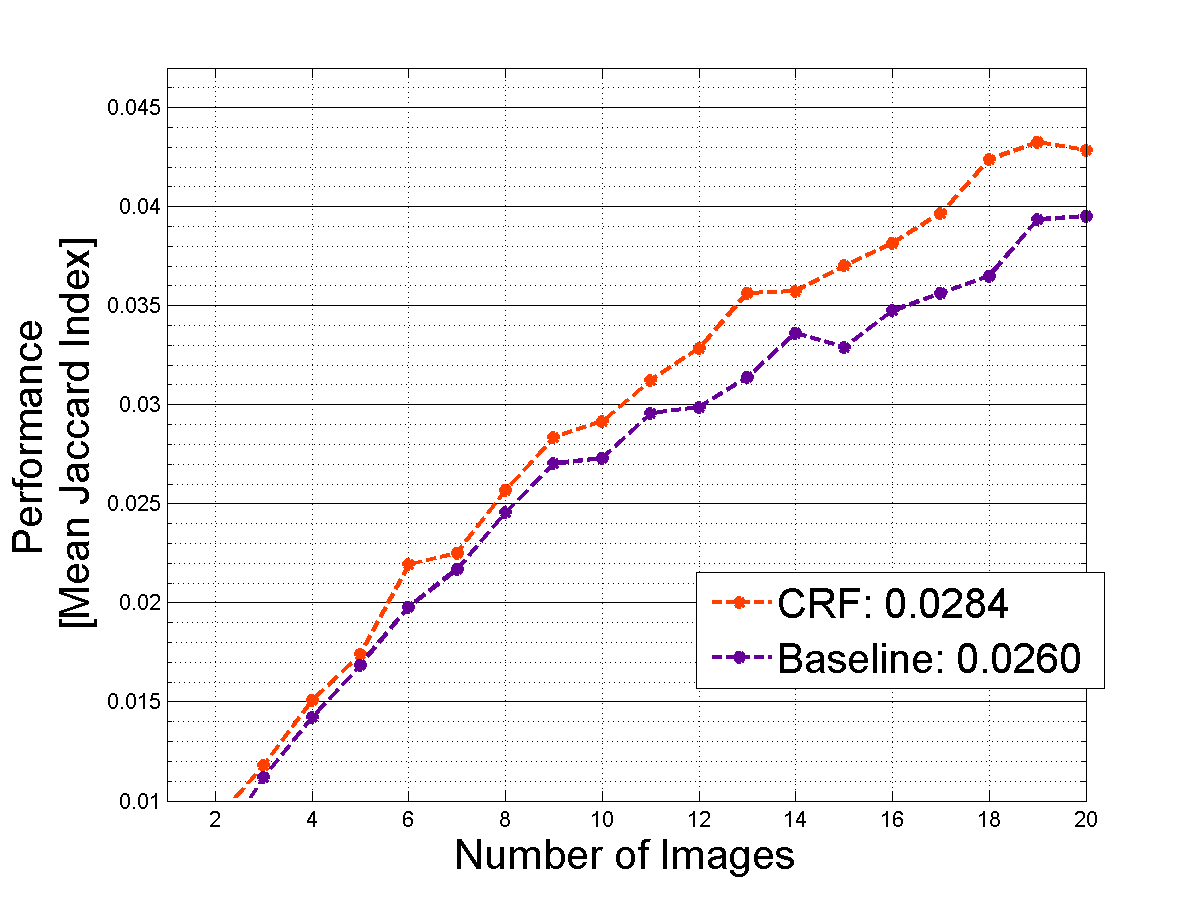} 
}

   \caption{\small{Evaluation of the presented algorithm to map a set of words from an open vocabulary to a dictionary. The output of the random walk algorithm is mapped using two algorithms: the baseline and the proposed CRF  algorithm. The baseline maps every output word to its closest word as it is measured by the cosine distance in the vector space of words. The CRF algorithm outperforms the baseline under different sorting conditions and the number of query images.}}
\label{fig:base3}
\end{figure*}

For the quantitative experiments, the performance is reported as the average Jaccard index of the 300 topics as the function of  the number of input images. The order in which images are presented has an impact on the performance, since some images are more significant than others in describing the topics. We use the provided human annotations in the groundtruth, that describe the relevance of an image for the topic, to generate three different sorting of the input images. They are: a) images preserve the original dataset order; b) Images are sorted in decreasing order, using the most descriptive images of the topic first, and c) Images are sorted in increasing order, using the least descriptive images of the topic first. As is evident from figures \ref{fig:base}, \ref{fig:base2}, and \ref{fig:base3} the order in which images are presented to the system has an impact on the performance. As is expected, better topic descriptions are achieved when a larger number of query images are used.

Figure \ref{fig:base} shows the results for the baseline method. Each plot contains results by varying the number of retrieved words that describe the topic from 5 to 20 in intervals of 5 words. For the three types of sorting, retrieving the top 10 words produce the best results according to the mean of the average Jaccard index for different number of input images, and it will be considered the baseline method.  Figure \ref{fig:base2} compares the results of our method after applying the random walk algorithm against the baseline. As is evident from the figure, the random walk algorithm significantly outperforms the baseline under any type of sorting, and the number of  input images. Finally, figure \ref{fig:base3}, evaluates the performance of the algorithm to map a set of words form an open vocabulary to words from a dictionary. We used as starting point the output of the random walk algorithm of the figure \ref{fig:base2}, which is mapped using a baseline mapping method and the CRF based mapping algorithm. The baseline maps every output word to its closest significant word in the dictionary according to the cosine distance in the vector space of words. The CRF algorithm preserves the semantic similarity under different sorting and amount of query images, outperforming the results of the baseline mapping method.



\section{Conclusions}
In this paper, we present the problem of identifying the topic that is being depicted by a group of images. We proposed a completely unsupervised approach that allows us to obtain a set of words that properly describe the topic. The approach relies on a combination of image retrieval, auto-tagging, and random walk that take into account the semantic similarity among the words. We also proposed a CRF based algorithm to map a set of words from an open vocabulary to a closed dictionary preserving the semantic similarity. The results indicates that the proposed algorithm clearly improves the proposed baselines.

\newpage
{\small
\bibliographystyle{ieee}
\bibliography{ICCV2015}
}
\end{document}